\documentclass[11pt,a4paper]{article}
\usepackage{authblk}
\usepackage[hyperref]{emnlp2020}
\usepackage{times}
\usepackage{latexsym}

\usepackage{microtype}

\aclfinalcopy 

\usepackage{graphicx}
\usepackage{url}
\usepackage{booktabs}

\usepackage[fleqn]{amsmath}
\usepackage{amssymb}
\usepackage{multirow}

\usepackage{xspace}

\newcommand{\model}{{Octa}\xspace}

\usepackage[normalem]{ulem}

\usepackage{nccmath}

\newcounter{sentence}
\usepackage[many]{tcolorbox}
\newtcolorbox{sentence}[2][]{
tikznode boxed title,
enhanced,
interior style={white},
boxsep=1pt,left=1pt,right=1pt,bottom=1pt,
width=\columnwidth,
boxrule=1pt,
attach boxed title to top center= {yshift=-\tcboxedtitleheight/2},
colbacktitle=white,coltitle=black,
boxed title style={size=normal,colframe=white,boxrule=0pt},
title={#2},
before title={\refstepcounter{sentence}},
before upper={\def\temp{#1}\ifx\temp\empty\else\textbf{#1.} \fi
 \def\@currentlabel{\p@sentence\thesentence}}
}

\newcommand{\myparagraph}[1]{\smallskip\noindent{\textbf{#1.}~}}

\newcommand{\fsc}{\textsc}
\newcommand{\fsl}{\textsl}

\newcommand*\samethanks[1][\value{footnote}]{\footnotemark[#1]}
\usepackage[leftcaption,raggedright]{sidecap}
\sidecaptionvpos{figure}{t}

\title{Octa: Omissions and Conflicts in Target-Aspect Sentiment Analysis}

\author[$\dagger$]{\textbf{Zhe Zhang} \thanks{~~Equal contribution.}\protect\phantom{\footnotesize 1}}
\author[$\dagger$]{\textbf{Chung-Wei Hang} \samethanks \protect\phantom{\footnotesize 1}}
\author[$\ddagger$]{\textbf{Munindar P.~Singh}}
\affil[$\dagger$]{IBM Corporation}
\affil[$\ddagger$]{Department~of Computer~Science, North~Carolina~State~University}
\affil[ ]{\tt {\{zhangzhe,hangc\}@us.ibm.com, singh@ncsu.edu}}

\date{}

\begin{document}
\maketitle

\begin{abstract}
Sentiments in opinionated text are often determined by both aspects and target words (or \emph{targets}). We observe that targets and aspects interrelate in subtle ways, often yielding conflicting sentiments. Thus, a naive aggregation of sentiments from aspects and targets treated separately, as in existing sentiment analysis models, impairs performance. 

We propose \model,\footnote{The data and source code of Octa can be found at \url{https://github.com/chungweihang/octa}} an approach that jointly considers aspects and targets when inferring sentiments. To capture and quantify relationships between targets and context words, \model uses a selective self-attention mechanism that handles implicit or missing targets. Specifically, \model involves two layers of attention mechanisms for, respectively, selective attention between targets and context words and attention over words based on aspects. On benchmark datasets, \model outperforms leading models by a large margin, yielding (absolute) gains in accuracy of 1.6\% to 4.3\%.
\end{abstract}

\section{Introduction}
\label{sec:introduction}
People share their opinions about almost anything: tourist attractions, restaurants, car dealerships, and products. Such opinionated texts do not merely help people make decisions in their daily life, but also help businesses measure consumer satisfaction to improve their offerings.

Sentiment analysis involves many aspects of Natural Language Processing, e.g., negation handling \cite{zhu-etal:2014}, entity recognition \cite{mitchell-etal:2013}, topic modeling \cite{zhang-s:limbic,zhangS:2019}. Importantly, opinionated texts often convey conflicting sentiments. Distinct sentiments may refer to distinct \emph{aspects} of the domain in question---e.g., food quality of a restaurant or battery life of a smartphone. These predefined domain aspects may or may not appear in the texts. \textbf{Aspect-Based Sentiment Analysis (ABSA)} approaches \cite{Wang:16,Li:18,liang-etal:19} predict sentiments from text about a given aspect. And, \textbf{Target-Based Sentiment Analysis (TBSA)} approaches \cite{ChenSBY:17,FanFZ:18,li-etal:18,du-etal:19,zhang-etal:19} predict sentiments of \emph{targets} that appear in an opinionated text. Targets are usually entities in a review: e.g., a dish for a restaurant and a salesperson for a car dealership. 

We posit that aspects and targets provide subtle, sometimes contradictory, information about sentiment and should therefore be modeled, not in isolation, but jointly. Considering them separately, as ABSA and TBSA approaches do, impairs performance. Take this review sentence from SemEval-15 as an example:

\begin{sentence}{Conflicting Sentiments on Aspect}
 \emph{We both had the \textbf{filet}, very good, didn't much like the \textbf{frites} that came with.}
\end{sentence}

If we ask about aspect \emph{Food\#Quality}, by disregarding targets during training, ABSA models fail to address the contradiction in sentiment about \emph{filet} and \emph{frites}, as do TBSA models, which focus on targets and disregard aspects. In the following review sentence from SemEval-16, the target \emph{fish} is associated with opposite sentiments: positive for \emph{Food\#Quality} and negative for \emph{Food\#Style\_options}.

\begin{sentence}{Conflicting Sentiments on Target}
\emph{The \textbf{fish} was fresh , though it was cut very thin.}
\end{sentence}

Opinionated text is often not structured. Users may not always mention targets explicitly. In some cases, the entities in a sentence are not the targets associated with the sentiment. In other cases, users mention multiple targets with sentiments in a sentence, but we need the overall sentiment. Consider the following two sentences from SemEval-16:

\begin{sentence}{Implicit or Missing Target}
(1) \emph{You are bound to have a very charming time.}
(2) \emph{Endless fun, awesome music, great staff!!!}
\end{sentence}

Here, (1) contains entity \emph{You} and positive sentiment toward aspect \emph{Restaurant\#General} but omits mention of the target \emph{restaurant}. And, (2) contains positive sentiment toward aspects \emph{Ambience} and \emph{Service}. It expresses a positive sentiment toward aspect \emph{Restaurant\#General} albeit with no target. How can we extract sentiments given an aspect with or without a target?

\paragraph{Contributions}
We propose \model, an approach that jointly considers aspects and targets. \model uses a selective attention mechanism to capture subtle Target-Context and Target-Target relationships that reduce noisy information from irrelevant relations. \model uses (1) aspect embeddings with attention to incorporate aspect dependencies and (2) a surrogate target with BERT sequence embeddings to handle implicit or missing targets. \model can classify different types of conflicting sentiments with aspects only, targets only, both, or none.

\model yields strong results on six benchmark datasets including SentiHood and four SemEval datasets, i.e., 2014 (target and aspect), 2015, and 2016. \model\ outperforms 16 state-of-the-art baselines by absolute gains in accuracy from 1.6\% to 4.3\%. 

\paragraph{Sample Results of \model}

We explain the benefit of \model via a few examples from the SemEval-16 test set in Table~\ref{tab:example}.
\begin{table*}[htb!]
\centering
\scalebox{0.93} {\footnotesize
\begin{tabular}{l l l l l}\toprule
&Sentence&Aspect&Target&Sent.\\\midrule
\multicolumn{1}{c}{\multirow{1}{0.1cm}[-.5em]{(a)}}&\multicolumn{1}{l}{\multirow{1}{*}[-.5em]{\emph{The \textbf{fish} was fresh , though it was cut very thin.}}}&\multicolumn{1}{l}{Food\#Quality} & \emph{fish}&\multicolumn{1}{l}{POS}\\
 & & Food\#Style\_options & \emph{fish} & NEG\\\midrule
\multicolumn{1}{c}{\multirow{3}{0.1cm}{(b)}}&\multicolumn{1}{l}{\multirow{3}{9cm}{\emph{\textbf{Food} wise, it's ok but a bit pricey for what you get considering the \textbf{restaurant} isn't a fancy place.}}}&\multicolumn{1}{l}{Food\#Quality}& \emph{Food} &\multicolumn{1}{l}{NEU}\\
 & & Restaurant\#Prices &\emph{restaurant} & NEG\\
 & & Ambience\#General & \emph{restaurant} & NEU\\\midrule
\multicolumn{1}{c}{\multirow{3}{0.1cm}{(c)}}&\multicolumn{1}{l}{\multirow{3}{9cm}{\emph{The \textbf{music} playing was very hip, 20-30 something pop music, but the \textbf{subwoofer to the sound system} was located under my seat, which became annoying midway through dinner.}}}&\multicolumn{1}{l}{\multirow{1}{*}[-.3em]{Ambience\#General}}& \multicolumn{1}{l}{\multirow{1}{*}[-.3em]{\emph{music}}} &\multicolumn{1}{l}{\multirow{1}{*}[-.3em]{POS}}\\
& &\multicolumn{1}{l}{\multirow{1}{*}[-.4em]{Ambience\#General }}& \multicolumn{1}{l}{\multirow{1}{*}[-.3em]{\emph{subwoofer to the}}} &\multicolumn{1}{l}{\multirow{1}{*}[-.4em]{NEG}} \\
&&&\multicolumn{1}{l}{\multirow{1}{*}[-.1em]{\emph{sound system}}}&
\\\midrule
\multicolumn{1}{c}{\multirow{1}{0.1cm}{(d)}}&\multicolumn{1}{l}{\multirow{1}{9.5cm}{\emph{As part of a small party of four, our food was dropped off without comment}}} &Service\#General&--- &NEG\\\midrule
\multicolumn{1}{c}{\multirow{3}{0.1cm}{(e)}}&\multicolumn{1}{l}{\multirow{3}{*}{\emph{Endless fun, awesome \textbf{music}, great \textbf{staff}}}}&\multicolumn{1}{l}{Ambience\#General}& \emph{music} &\multicolumn{1}{l}{POS}\\
 & & Service\#General & \emph{staff} & POS\\
 & & Restaurant\#General & ---  & POS\\
\bottomrule
\end{tabular}
}
\caption{Sample results of \model.}
\label{tab:example}
\end{table*}
In case (a), the \textbf{same target} is paired with \textbf{different aspects}. \model detects positive sentiment toward aspect \emph{Food\#Quality} based on target \emph{fish} and context \emph{fresh}. By attending to different context \emph{cut very thin} but the same target, \model detects negative sentiment toward aspect \emph{Food\#Style\_options}. In case (b) where \textbf{different aspects} paired with the \textbf{same or different targets}, \model correctly detects neutral sentiment toward target \emph{food} for  aspect \emph{Food\#Quality}. For target \emph{restaurant}, \model successfully detects conflicting sentiments toward different aspects by 
locating different context words. In case (c), the \textbf{same aspects} are paired with \textbf{different targets}. \model correctly detects the conflicting sentiments toward the same aspect \emph{Ambience\#General}. Case (d) has \textbf{aspect with implicit target} and case (e) has \textbf{different aspects with or without target}. \model successfully detects the sentiment toward implicit or missing target.

\section{Problem Definition}

The input of our sentiment analysis task is a sequence of words, with an aspect, or a target, or both. Our goal is to identify the sentiment polarity associated with the aspect and the target. Formally, \model has three inputs,

\begin{itemize}
 \item Sequence of words: $W=\{w_1,\ldots, w_N\}$,
 \item Target $T_i=\{t_1, \ldots, t_M\}$ where $t_i \in W$, and
 \item Aspect $a_i \in A=\{a_1,\ldots,a_{|A|}\}$ where $A$ is a set of aspects.
\end{itemize}

The remaining words that are not part of the target are context words $C=\{c_1, \ldots, c_{N-M}\}$.

\section{\model Model Overview}

\begin{figure}
 \centering
 \includegraphics[width=0.99\linewidth]{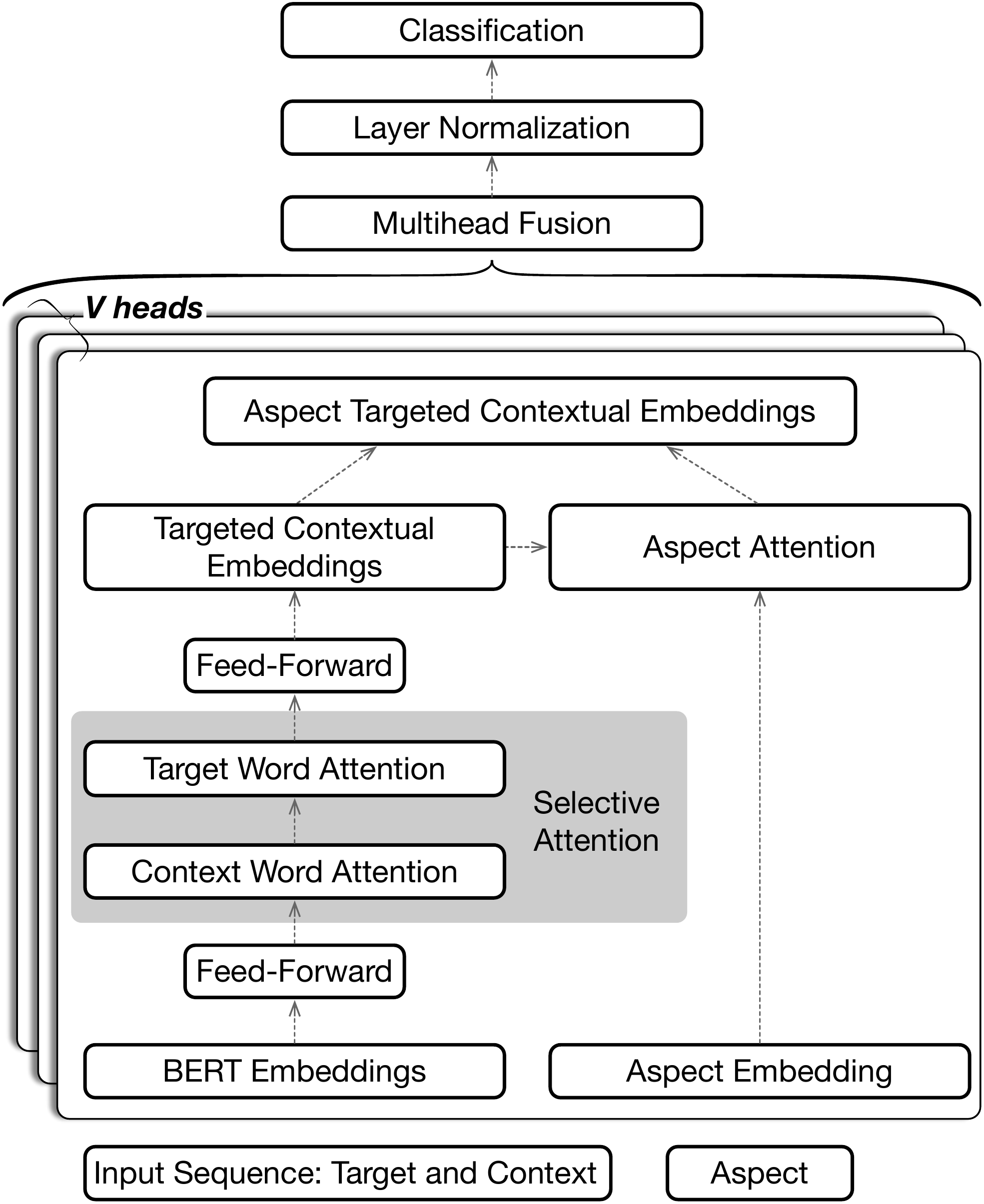}
 \caption{Architecture of \model.}
 \label{fig:overall}
\end{figure}

Figure~\ref{fig:overall} shows the \model architecture. To infer the sentiment for an aspect and a target composed of words from the sequence, first, \model uses BERT to generate word embeddings. Second, \model uses a selective attention mechanism to compute context word and target attention weights and applies them to word embeddings to generate targeted contextual embeddings. Third, \model constructs aspect embeddings and uses the embeddings to compute aspect attention over target and context words.

\model uses a multihead architecture to learn attention in diverse embedding subspaces. It fuses and normalizes embeddings from each head and uses a linear classification layer with a softmax activation for sentiment classification. To introduce nonlinearity, \model uses feed-forward networks (shown in grey), each comprising two fully connected layers followed by a nonlinear activation. 

\subsection{BERT Embeddings}

\model uses Bidirectional Encoder Representations from Transformers (BERT) \cite{bert:19} to generate word embeddings. BERT is a contextualized language representation model, pretrained on large corpora and fine-tuned on downstream tasks, including token-level classification (named entity recognition and reading comprehension) and sequence-level classification (semantic similarity and sentiment analysis). Despite its success on various benchmarks, BERT ignores the relationships among target words, context words, and aspects, which are crucial for sentiment analysis.

\subsection{Selective Self-Attention Mechanism}
Words connect with one another to form semantic relations and create meanings in different contexts. Self-attention \cite{Vaswani:17} seeks to quantify this process. To capture relationships between words, it learns to represent each word using itself and the other words in the same sentence. The flexible structure of self-attention provides benefits in capturing different relations without range restriction. An ideal self-attention layer should attend to relations differently to create contexts for different goals. In practice, such flexibility may introduce noisy relations that lead to less-focused attention and confuse the decision layer.

In opinionated texts, context words carry sentiment. A context word can be associated with one or more targets. Thus, capturing Target-Context relationships is pivotal. We posit that capturing Target-Target relationships is important when targets contain multiple words. Context words can carry different sentiment when the same target word paired with other target words. For example, in the sentences \emph{The wine list is long} and \emph{The waiting list is long}, context word, \emph{long}, is positive for target \emph{wine list} but negative for target \emph{waiting list}. 

\begin{figure*}[htb]
 \centering
 \includegraphics[width=1\textwidth]{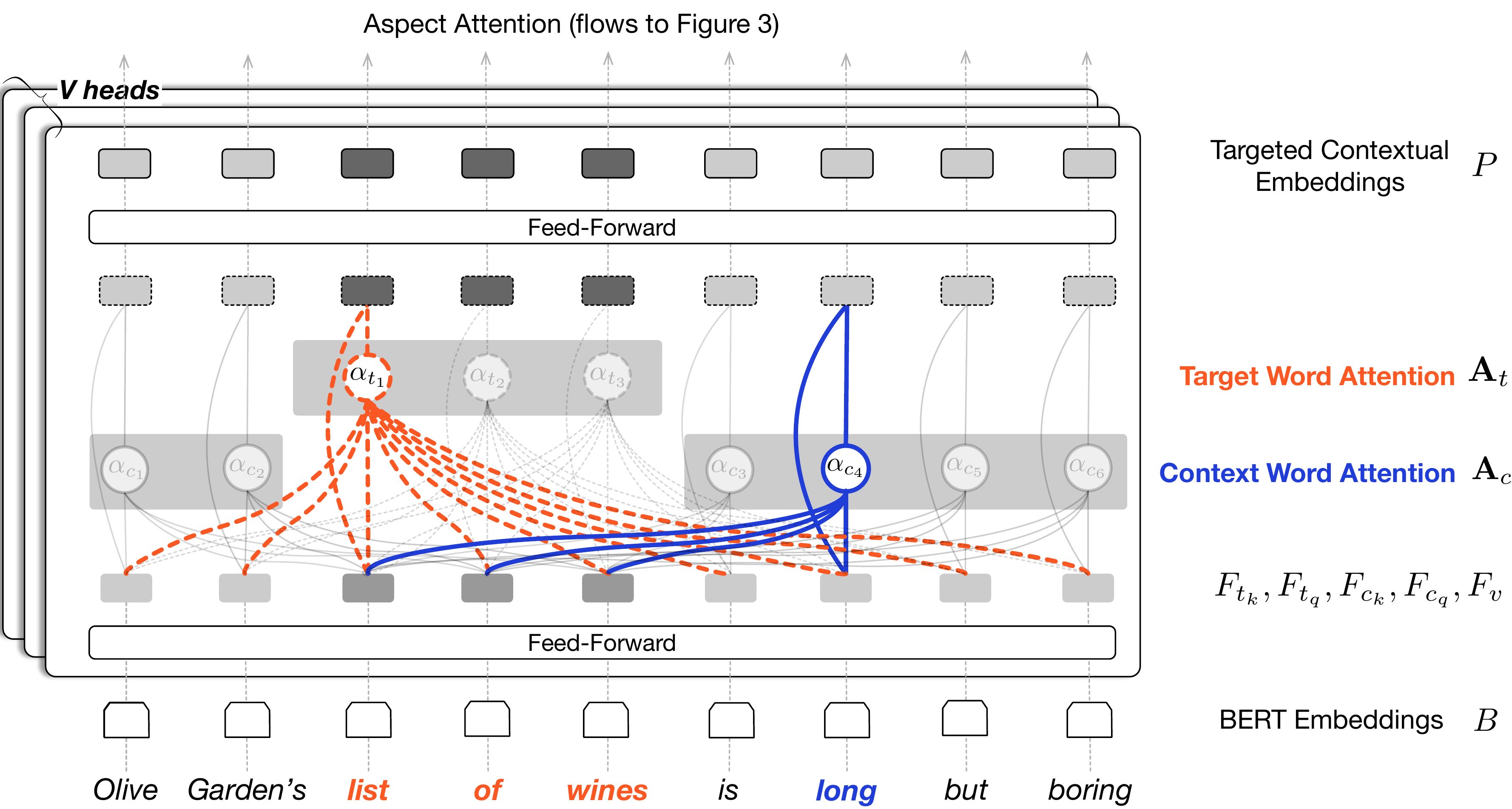}
 \caption{An illustration of target attention.}
 \label{fig:target-attention}
\end{figure*}

\model uses a selective self-attention encoder to capture the subtle Target-Context and Target-Target relationships. Figure~\ref{fig:target-attention} shows the encoding process.

Formally, given a sentence containing one target $t$ that consists of $M$ target words and context $c$ that consists of $N$ context words, let $B_t=[b_{t_1},\ldots, b_{t_M}] \in \mathbb{R}^{M\times d_{B}}$, $B_c=[b_{c_1},\ldots, b_{c_N}]\in \mathbb{R}^{N\times d_{B}}$ denote the BERT embedding matrices of targets and context words, respectively, where $d_{B}$ is the dimension of BERT embeddings. We use BERT's {\tt [CLS]} token as either a target (when no target is provided) or a context word.

\myparagraph{Feed-Forward Networks} \model adopts a key-query-value attention structure \cite{Vaswani:17} where keys, queries, and values are projected vectors. The structure first combines each query with all of keys through a compatibility function to generate attention weights. Then, it uses the weights to combine corresponding values to generate the output. \model uses five feed-forward networks to construct keys, queries, and values for target and context words. Each feed-forward network comprises two fully connected linear layers connected by a GELU \cite{hendrycks:2016} activation for element-wise nonlinear projection. 
\begin{align}
\label{eq:feedforwardtk}
F_{t_k} &= W_{t_{k1}}\cdot(\text{GELU}(W_{t_{k2}}\cdot B_t)),\\
\label{eq:feedforwardtq}
F_{t_q} &= W_{t_{q1}}\cdot(\text{GELU}(W_{t_{q2}}\cdot B_t)),\\
\label{eq:feedforwardck}
F_{c_k} &= W_{c_{k1}}\cdot(\text{GELU}(W_{c_{k2}}\cdot B_c)),\\
\label{eq:feedforwardcq}
F_{c_q} &= W_{c_{q1}}\cdot(\text{GELU}(W_{c_{q2}}\cdot B_c)),\\
\label{eq:feedforwardv}
F_v &= W_{v_1}\cdot(\text{GELU}(W_{v_2}\cdot [B_t\oplus B_c])),
\end{align}
where $F_{t_k}, F_{t_q} \in \mathbb{R}^{M\times d_F}$ are keys and queries of targets, $F_{c_k}, F_{c_q} \in \mathbb{R}^{N\times d_F}$ are keys and queries of context words, $F_v \in \mathbb{R}^{(M+N)\times d_F}$ are values for both kinds of words, $\oplus$ means matrix vertical concatenation, $W_{(\cdot)}$ are parameters to learn, and we omit the bias for simplicity. 

\myparagraph{Target Word Attention} \model constructs an affinity matrix $\mathbf{A}_t=\{\alpha_{t_1}, \ldots, \alpha_{t_M}\} \in \mathbb{R}^{M\times(M+N)}$ by computing dot products of each target with each word in the sentence. 
\begin{align}
\label{eq:targetaffinity}
\mathbf{A}_t=\text{softmax}(F_{t_q} \cdot [F_{t_k}\oplus F_{c_k}]^T).
\end{align}
$\mathbf{A}_t$ is normalized row-wise to generate a list of attention weights for each target. These attention weights quantify relations between words and describe the amount of focus the encoder should place on other words when encoding a target. For sentences with no target, \model uses BERT's {\tt [CLS]} token as a surrogate target to leverage the aggregated sentence information.

\myparagraph{Context Word Attention} \model creates a mask matrix $\mathbf{K}_c=\{k_{c_1}, \ldots, k_{c_N}\} \in \mathbb{R}^{N\times (M+N)}$. Here, $k_{c_i}$ equals 1.0 if the corresponding position is context word $c_i$ or a target and zero otherwise. \model constructs the affinity matrix $\mathbf{A}_c=\{\alpha_{c_1}, \ldots, \alpha_{c_N}\} \in \mathbb{R}^{N\times (M+N)}$ by computing the dot products of each context word with itself and each target in the sentence masked by $\mathbf{K}_c$, where $\circ$ denotes Hadamard product.
\begin{align}
\label{eq:contextaffinity}
\mathbf{A}_c=\text{softmax}(F_{c_q}\! \cdot [F_{t_k}\oplus F_{c_k}]^T\!\circ \mathbf{K}_c).
\end{align}
$\mathbf{A}_c$ is normalized row-wise to generate a list of attention weights for each context word. These attention weights quantify dependencies between each context word and each target. Our mask removes noisy dependencies between the context words.

\myparagraph{Targeted Contextual Embeddings} Given target attention $\mathbf{A}_t$ and context word attention $\mathbf{A}_c$, \model computes targeted contextual embeddings $P \in \mathbb{R}^{(M+N)\times d_F}$ as follows.
\begin{align}
\label{eq:contextembeddings}
P=[\mathbf{A}_t\oplus \mathbf{A}_c] \cdot F_v.
\end{align}

\subsection{Aspect Attention}

How the aspects and words in a sentence relate is vital in inferring sentiments. As the second review sentence in Section~\ref{sec:introduction} shows, one target can associate with different sentiments for different aspects.

To incorporate aspect information, given $L$ aspects, $A=\{a_{1},\ldots, a_{L}\}$, \model learns a list of aspect embeddings $F_{A}=\{f_{a_1},\ldots, f_{a_L}\} \in \mathbb{R}^{L\times{d_E}}$ as follows,
\begin{align}
\label{eq:feedforwardA}
F_{A}=W_{A_{1}}\cdot(\text{GELU}(W_{A_{2}}\cdot E)),
\end{align}
where $E=\{e_{a_1}, \ldots, e_{a_L}\}$, $e_{a_i} \in \mathbb{R}^{d_E}$ are a list of randomly initialized aspect keys, $W_{A_1}$ and $W_{A_2}$ are weights to learn, and bias is omitted for simplicity.  To capture the relationships, \model builds the affinity matrix $\mathbf{A}_a \in \mathbb{R}^{M+N}$ between aspect embeddings $f_{a_i}$ and targeted contextual embeddings $P$ of the sentence. 

\begin{figure*}[htb]
 \centering
 \includegraphics[width=1\textwidth]{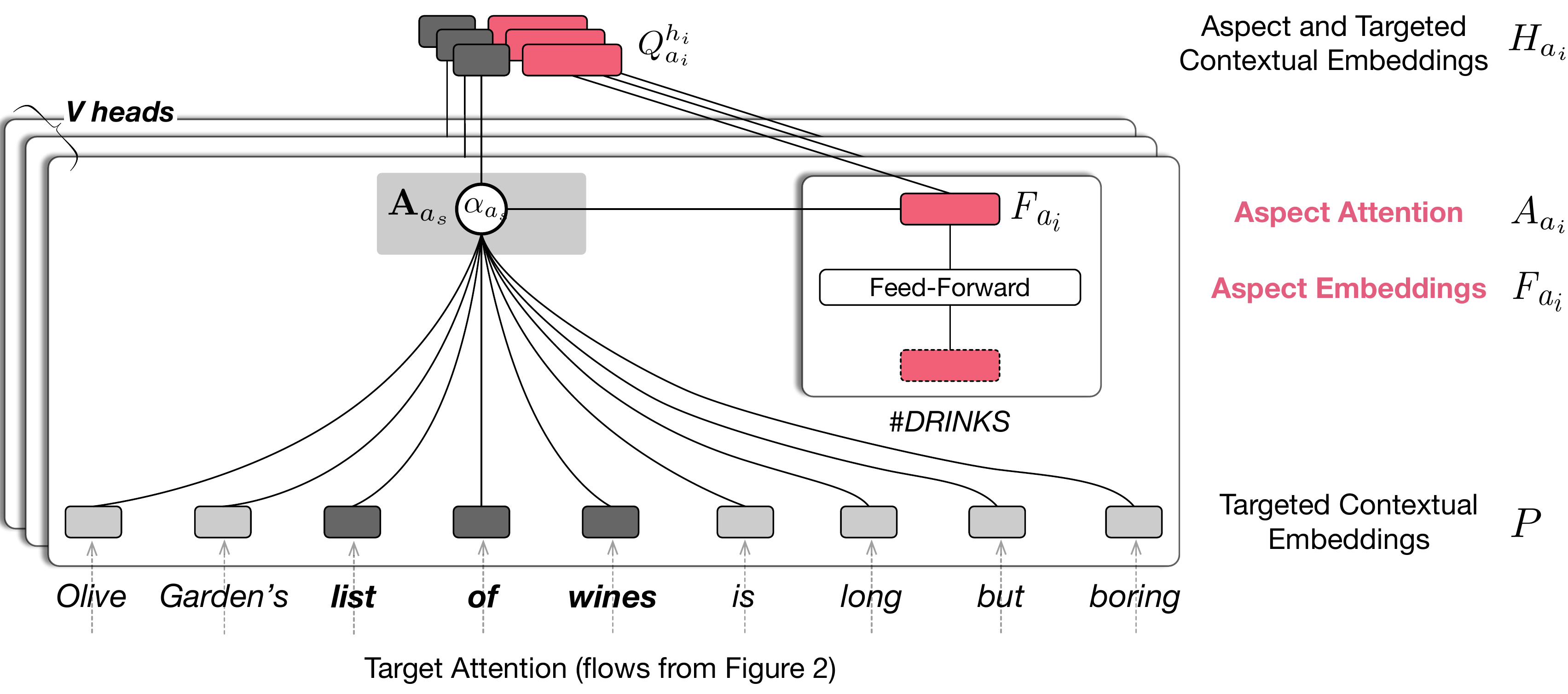}
 \caption{An illustration of aspect attention.}
 \label{fig:aspect-attention}
\end{figure*}

An illustrative example of aspect attention is shown in Figure~\ref{fig:aspect-attention}.
\begin{align}
\label{eq:aspectaffinity}
\mathbf{A}_{a_i}=\text{softmax}(f_{a_i} \cdot P^T).
\end{align}
The aspect and targeted contextual embeddings $Q_{a_{i}}$ for aspect $a_i$, $Q_{a_{i}} \in \mathbb{R}^{d_E+d_F}$, are computed as
\begin{align}
\label{eq:aspectcontextembeddings}
Q_{a_{i}}= [\mathbf{A}_{a_i} \cdot P] \odot f_{a_i},
\end{align}
where $\odot$ denotes horizontal matrix concatenation.

\subsection{Multihead Fusion}

To attend in parallel to relation information from different dimensional subspaces, \model uses a multihead architecture with $V$ heads. The final aspect and targeted contextual embeddings $H_{a_{i}} \in \mathbb{R}^{V *(d_E+d_F)} $for aspect $a_i$ is the fusion of all heads.

\begin{align}
\label{eq:multihead}
H_{a_{i}}= [Q^{h_1}_{a_{i}}\odot, \ldots, \odot~ Q^{h_V}_{a_{i}}].
\end{align}

\subsection{Sentiment Classification}

For sentiment classification, \model first applies layer normalization \cite{BaKH:16} on the multihead fusion. Then, it uses a fully connected linear layer followed by a softmax activation to to project $H_{a_{i}}$ to $y\in \mathbb{R}^S$, the posterior probability over $S$ sentiment polarities, is $y$ (omitting the bias):

\begin{align}
\label{eq:final}
y= \text{softmax}(W_y \cdot H_{a_{i}}),
\end{align}
where $W_y$ is parameter to learn. We train \model with cross-entropy  loss.

\section{Empirical Evaluation}

\subsection{Data}

We train and evaluate \model on six benchmark datasets, described in Table~\ref{tab:datasets}, from three domains.

\begin{SCtable*}
\centering
\begin{tabular}{l@{~} c@{~~} c@{~~} r@{~~} r@{~~} r@{~~} r@{~~} r@{~~} r}\toprule
\multicolumn{1}{c}{\multirow{1}{*}[-.5em]{Dataset}}&&&\multicolumn{2}{c}{Positive}&\multicolumn{2}{c}{Neutral} &\multicolumn{2}{c}{Negative}\\
 & Aspects & Labels & Train & Test& Train & Test & Train & Test\\\midrule
SemEval-14A & 5 & 3 & 2,179 & 657 & 695 & 146 & 839 & 222\\
SemEval-14T & 5 & 3 & 2,164 & 728 & 724 & 210 & 805 & 196 \\
SemEval-15 & 13 & 3 & 1,198 & 454 & 53 & 45 & 403 & 346\\
SemEval-16 & 12 & 3 & 1,657 & 611 & 101 & 44 & 749 & 204\\
SentiHood-dev & 12 & 2 & 2,480 & 616 & -- & -- & 921 & 224 \\
SentiHood-test & 12 & 2 & 2,480 & 1,217 & -- & -- & 921 & 462 \\
\bottomrule
\end{tabular}
\caption{Datasets. SemEval has restaurant review sentences. SentiHood has sentences about urban neighborhoods. SemEval-14-T has sentiments for targets without aspects and SemEval-14-A for aspects without targets. SemEval-15, SemEval-16, and SentiHood have targets and aspects.}
\label{tab:datasets}
\end{SCtable*}

\subsection{Parameter Settings}

We set the dimension of aspect embeddings $d^E$ to 1,024. For all feed-forward networks, we use 1,024 as the dimension of both inner and outer states $d^F$.
We train \model with 16 attention heads and freeze aspect embeddings during training. 

We follow the literature in that we do not further split SemEval training sets into training and validation sets due to their size. Instead, we use SentiHood-dev for parameter tuning. For regularization, we add dropouts with a rate of 0.1 between the two fully connected layers in each nonlinear feed-forward network. For optimization, we use Adam \cite{KingmaB:14} and set $\beta_{1}=0.9$, $\beta_{2}=0.99$, weight decay = 0.01, and the learning rate = 1e-5, with a warmup over 0.1\% of training. 

For all experiments, we train \model for 10 epochs on mini-batches of 32 randomly sampled sequences of 128 tokens. We repeat the training and testing cycle five times using different random seeds. Our evaluation metrics include accuracy and macro F$_1$ score. We perform the two-sampled t-test on the improvement of \model over BERT. As reported in \cite{bert:19}, we observe unstable performance for both \model and BERT. We perform several restarts and select best performed models. For model size, \model introduces 2.5\% more parameters (343M) compared with BERT sequence classification (335M, whole word masking). Training on SemEval-16 with single NVIDIA Tesla V100 takes 69 seconds/epoch for \model and 65 seconds/epoch for BERT.

\subsection{Baselines}

We compare the performance of \model against the following published models.

\paragraph{Feature based Baselines:}

\textbf{NRC-Canada}, \textbf{DCU}, \textbf{Sentiue}, and \textbf{XRCE} require feature engineering based on linguistic tools and external resources. Of these, NRC-Canada and DCU achieve the best performance on SemEval 2014 sentiment classification for aspect category and aspect term, respectively. Sentiue and XRCE are the best performing for SemEval 2015 and 2016, respectively.

\paragraph{TBSA Baselines:} \textbf{RAM} \cite{ChenSBY:17} builds position-weighted memory using two stacked BiLSTMs and the relative distance of each word to the left or right boundary of each target. It uses a GRU with multiple attention computed using the memory. \textbf{TNet-AS} \cite{li-etal:18} dynamically associates targets with sentence words to generate target specific word representation and uses adaptive scaling to preserve context information. \textbf{MGAN} \cite{FanFZ:18} is an attention network based on BiLSTM that computes coarse-grained attention using averaged target embeddings and context words and leverages word similarity to build fine-grained attention. \textbf{IACapsNet} \cite{du-etal:19} leverages capsule network to construct vector-based feature representation. It uses interactive attention EM-based capsule routing mechanism to learn the semantic relationship between targets and context words. \textbf{TNet-ATT} \cite{tang:2019} leverages the relation between context words and model's prediction as supervision information to progressively refine its attention module for aspect based sentiment classification. \textbf{ASGCN-DG} \cite{zhang-etal:19} builds Graph Convolutional Networks over dependency trees and uses masking and attention mechanisms to generate aspect-oriented sentence representations. \textbf{TD-GAT-BERT} \cite{huang-carley} uses a Graph Attention Network to capture dependency relationship among words and an LSTM to model target related information.

\paragraph{ABSA Baselines:} \textbf{ATAE-LSTM} \cite{Wang:16} is based on LSTM. It uses aspect embeddings to learn attention weights. \textbf{GCAE} \cite{Li:18} is a CNN with two convolutional layers that use different nonlinear gating units to extract aspect-specific information. \textbf{AGDT} \cite{liang-etal:19} contains an aspect-guided encoder which consists of an aspect-guided GRU and a deep transition GRU to extract aspect-specific sentence representation. Note that GCAE and AGDT can be  extended for TBSA. However, neither of them 
jointly considers both aspects and targets and therefore fails to handle conflicting sentiments.

\paragraph{Other Baselines:} \textbf{Sentic LSTM} \cite{MaPC:18} uses an LSTM with a hierarchical attention mechanism to model both target and aspect attention. It incorporates commonsense knowledge into sentence embeddings. \textbf{BERT} does not consider aspects and targets. We compare with BERT to evaluate the performance gain from selective attention.  We use the whole world masking pretrained BERT in our experiments.  Additional results using BERT base and large models are in Appendix~\ref{sec:appendix}.

\subsection{Results}
\label{sec:results}

\newcommand{\sd}{\text{, sd }}
\newcommand{\pv}{\text{, p }}

\begin{table*}[htb]
\centering
\scalebox{.88} {
\begin{tabular}{c@{~}c@{~~~~} l@{\hspace{2\tabcolsep}}r@{~~} r@{\hspace{4\tabcolsep}} r@{\hspace{3\tabcolsep}} r@{\hspace{4\tabcolsep}} r@{~~~~~~} r@{\hspace{4\tabcolsep}} r@{\hspace{2\tabcolsep}} r@{}}\toprule
&& \multirow{2}{*}{Model} & \multicolumn{2}{l}{{SemEval-14A~~~~~~~~}} & \multicolumn{2}{l}{{SemEval-14T}} & \multicolumn{2}{l}{{SemEval-15}} & \multicolumn{2}{l}{{SemEval-16}} \\\cmidrule{4-11}

&&& Acc. & F$_1$ & Acc. & F$_1$ & Acc. & F$_1$ & Acc. & F$_1$ \\
\midrule
\multirow{4}{*}{\scalebox{1}{\rotatebox[origin=c]{90}{Feature}}} &
\multirow{4}{*}{\scalebox{1}{\rotatebox[origin=c]{90}{Based}}} &NRC-Canada$^\sharp$ & 82.92& --& 80.05& -- & --&--& --&--\\
&& DCU$^\sharp$ & --& -- & 80.95 & -- & --&-- & -- & --\\
&& Sentiue$^\sharp$ & --& --& --& -- & 78.69 & -- & -- & -- \\
&& XRCE$^\sharp$ & -- & -- & -- & -- & -- & -- & 88.13 & -- \\
\midrule
\multirow{12}{*}{\scalebox{1}{\rotatebox[origin=c]{90}{Deep-Learning}}} &\multirow{12}{*}{\scalebox{1}{\rotatebox[origin=c]{90}{Based}}} 
& ATAE-LSTM$^\sharp$ & 77.20 &--& --&-- & --&--&--&--\\
&& GCAE$^\sharp$ & 79.35&-- & 77.28&-- & --&--&--&--\\
&& AGDT$^\sharp$ & 81.78&-- & 78.85&-- & --&--&--&--\\

&& RAM$^\natural$ & -- &--& 79.79&68.86 & --&--&--&--\\
&& MGAN$^\sharp$ & --&-- & 81.25&71.94 & --&--&--&--\\
&& TNet-AS$^\sharp$ & -- &--& 80.69&71.27 & --&--&--&--\\
&& IACapsNet$^\sharp$ & --&-- & 81.79&73.40 & --&-- & --&--\\
&& TNet-ATT$^\sharp$ & -- & -- & 81.53 & 72.90 & -- & -- & -- & --\\
&& ASGCN-DG$^\sharp$ & --&-- & 80.77&72.02 & 79.89& 61.89 & 88.99 & 67.48\\
&& Sentic LSTM$^\sharp$ & --&-- & --&-- & 76.47&-- & --&--\\
&& TD-GAT-BERT$^\sharp$ & --&-- & 83.00&-- & --&-- & --&--\\
&& BERT & \textbf{86.15}&78.70 & 80.39&{69.00} & 83.72&{65.63} & 88.52 & {74.68} \\
&& \textbf{\model} & {86.03} &\textbf{78.88} & \textbf{84.90}\rlap{\textsuperscript{\textdagger}} &\textbf{77.57}\rlap{\textsuperscript{\textdagger}} & \textbf{86.27}\rlap{\textsuperscript{\textdagger}} &\textbf{67.17} & \textbf{90.10}\rlap{\textsuperscript{\textdagger}} &\textbf{76.51} \\
&& p-value vs. BERT & 0.64 & 0.72 & 1.16e-6 & 1.27e-6 & 9.21e-4 & 8.14e-2 & 1.64e-4 & 6.12e-2 \\
\bottomrule
\end{tabular}
}
\caption{Comparing accuracy and F$_1$ on SemEval tasks. Note that only Sentic LSTM and \model can jointly consider aspects and targets. Results with $\sharp$ are obtained from the original papers. Results with $\natural$ are obtained from \cite{li-etal:18}. Throughout, $\ast$ and $\dagger$ indicate if performance of BERT is significantly different from that of \model at the levels of 0.05 and 0.001, respectively, measured by the two-sample $t$-test (p-values for the comparison with BERT are listed at the last row). See Appendix~\ref{sec:appendix} for additional significance test results.}
\label{tab:semeval}
\end{table*}

Table~\ref{tab:semeval} compares \model with baselines on SemEval datasets. For SemEval-14-A, AGDT outperforms GCAE, demonstrating the benefits of aspect-guided sentence representation. \model outperforms AGDT and NRC-Canada with accuracy gains of 4.3\% and 3.1\%, respectively. Since SemEval-14-A lacks target information, \model uses the BERT {\tt [CLS]} token as the target. The result shows the benefit of selective attention to capture implicit target information. 
\model and BERT yield comparable performance. We find that SemEval-14-A contains sentences with conflicting sentiments toward the same aspect. In the testing split, of 146 sentences labeled \fsc{neu}, 52 sentences show conflicting sentiments---e.g., ``the falafal was rather over cooked and dried but the chicken was fine'' is labeled \fsc{neu} for aspect \fsl{food} but contains positive sentiment toward target \fsl{chicken} and negative sentiment toward target \fsl{falafal}. We conjecture that such data defects undermine the benefit of selective attention.

SemEval-14-T lacks aspect labels so \model treats it as one aspect. \model outperforms all baselines with an accuracy gain of 1.9\% compared with the best performing baseline, TD-GAT-BERT, of 4.0\% over the feature-based baseline DCU.

SemEval-15 and SemEval-16 associate sentiment with both aspect and targets. \model outperforms all baselines. Specifically, \model obtains a 2.6\% and 1.6\% accuracy improvement over BERT on SemEval-15 and SemEval-16, respectively. The F$_1$ improvements over BERT are 1.5\% and 1.8\%. Also, \model outperforms the top feature-based models, Sentiue and XRCE. The results demonstrate the benefit of jointly considering aspects and targets.


\begin{table}[htb]
\centering

\scalebox{0.88} {
\begin{tabular}{l@{} rrrr}\toprule
\multicolumn{1}{c}{\multirow{2}{*}{Model}} & \multicolumn{2}{c}{{SentiHood-D}} & \multicolumn{2}{c}{{SentiHood-T}}\\\cmidrule{2-5}
& Acc. & F$_1$& Acc. & F$_1$\\ \midrule
Sentic LSTM &88.80&--&89.32&--\\
BERT &87.60 & 83.76 &87.09 & 83.02\\
\textbf{\model} & \textbf{92.17}\rlap{\textsuperscript{\textdagger}} & \textbf{89.86}\rlap{\textsuperscript{\textdagger}} & \textbf{91.34}\rlap{\textsuperscript{\textdagger}} & \textbf{89.00}\rlap{\textsuperscript{\textdagger}}\\
p-value & 8.32e-9 & 1.46e-8 & 5.08e-9 & 1.16e-8 \\
\bottomrule
\end{tabular}
}
\caption{Comparing performance on SentiHood data.}
\label{tab:sentihood}
\end{table}

Table~\ref{tab:sentihood} shows the results on SentiHood. \model outperforms the state-of-the-art Sentic LSTM with accuracy gains of 3.3\% and 2.0\% on dev and test, respectively. Sentic LSTM jointly considers both aspects and targets through a hierarchical attention mechanism. We attribute \model's performance to its nonrecurrent architecture, which alleviates the dependency range restriction in LSTM, and to its selective attention mechanism, which reduces noisy dependency information from irrelevant relations.


To further evaluate \model's capability of handling sentences with conflicting sentiments, we apply trained BERT and \model only on the conflicting samples from SemEval-15, SemEval-16, and SentiHood-test. There are 152, 96, 343 conflicting samples in SemEval-15, SemEval-16, and SentiHood-test, respectively. Table~\ref{tab:conflicting-samples} shows the results. We see that for all datasets, \model outperforms BERT with a large margin. The accuracy gains are 15.3\%, 11.5\%, and 19.8\%, respectively. 


\begin{table}[htb]
\centering
\scalebox{0.78} {
\begin{tabular}{l@{~~} r@{~~} r@{~~} r@{~~} r@{~~} r@{~~} r}\toprule
Model & \multicolumn{2}{c}{{SemEval-15}} & \multicolumn{2}{c}{{SemEval-16}} & \multicolumn{2}{c}{{SentiHood-T}} \\\midrule
& Acc. & $\text{F}_1$& Acc. & $\text{F}_1$ & Acc. & $\text{F}_1$ \\ \midrule
BERT & 52.55 & 39.26 & 52.50 & 43.88 & 53.24& 51.14\\
\textbf{\model} & \textbf{67.84}\rlap{\textsuperscript{\textdagger}} & \textbf{51.52}\rlap{\textsuperscript{\textdagger}} & \textbf{63.96}\rlap{\textsuperscript{*}} & \textbf{54.71} & \textbf{73.00}\rlap{\textsuperscript{\textdagger}} & \textbf{72.68}\rlap{\textsuperscript{\textdagger}}\\
p-value & 1.84e-8 & 7.74e-7 & 9.97e-3 & 6.12e-2 & 1.52e-7 & 2.47e-7 \\
\bottomrule
\end{tabular}
}
\caption{Comparing performance on conflicts.}
\label{tab:conflicting-samples}
\end{table}
\vspace{-0.4cm}
\subsection{Ablation Study}
We evaluate variants of \model on SemEval-15 to understand the contribution of aspects, targets, and selective attention. The same conclusion holds for the other datasets. As Table~\ref{tab:ablation} shows, using target selective attention (\model-Sel) yields 1.1\% better accuracy but similar $\text{F}_1$ as using aspect attention (\model-Asp). Combining aspect attention with target self-attention (\model-Asp-Full) hurts performance and stability, as seen in the lower accuracy and $\text{F}_1$, indicating that simply applying self-attention on targets and context words introduces noisy information. Replacing self-attention with selective attention (\model) yields gains in accuracy and $\text{F}_1$ of 4.3\% and 6.1\% respectively, indicating that selective attention is effective in combating noise.
\begin{table}[htb]
\centering
\scalebox{.87} {\begin{tabular}{lccrr}\toprule
Model & Aspect & Target & Acc. & $\text{F}_1$\\\midrule
\model-Asp & Yes & -- &84.09\rlap{\textsuperscript{\textdagger}} & 65.20\rlap{\textsuperscript{*}} \\
\model-Sel & -- & Selective &85.16 & 65.21 \\
\model-Asp-Full & Yes & Self &82.01\rlap{\textsuperscript{*}} & 61.08 \\
\textbf{\model} & \textbf{Yes} & \textbf{Selective} &\textbf{86.27} & \textbf{67.17} \\
\bottomrule
\end{tabular}
}
\caption{Comparing model variants on SemEval-15.}
\label{tab:ablation}
\end{table}

\section{Related Work}
Sentiment analysis has received substantial attention over the last few years. We highlight here only the works most relevant to \model.

\subsection{Aspect-Based Sentiment Analysis (ABSA)}

For the ABSA task, \citet{Wang:16} concatenate aspect embeddings with LSTM hidden states and apply attention mechanism to focus on different parts of a sentence given different aspects. \citet{Li:18} extracts features from text using a convolutional layer and propagates the features to a max pooling layer based on either aspects or targets. \citet{liang-etal:19} uses an aspect-guided encoder with an aspect-reconstruction step to generate either aspect- or target-specific sentence representation. The above models do not jointly consider aspects and targets and suffer when a target has conflicting sentiments toward different aspects. 

\subsection{Target-Based Sentiment Analysis (TBSA)}

For TBSA task, \citet{Tang:16} concatenate target and context word embeddings and use two LSTM models to capture a target's preceding and following contexts. \citet{ChenSBY:17} builds position-weighted memory using two stacked BiLSTMs and the relative distance of each word to the left or right boundary of each target. \citet{li-etal:18} dynamically associates targets with sentence words to generate target specific word representation and uses adaptive scaling to preserve context information. \citet{MajumderPGACE:18} uses a GRU with attention to generate an aspect-aware sentence representation and a multihop memory network to capture aspect dependencies. \citet{FanFZ:18} uses BiLSTM with attention mechanism to computes coarse-grained attention using averaged target embeddings and context words. It leverages word similarity to build fine-grained attention. \citet{Xu+19} prepend target tokens to a given text sequence, and predict sentiment based on BERT sequence embeddings. \citet{du-etal:19} leverages capsule network and uses interactive attention capsule routing mechanism to learn the relationship between targets and context words. 

\section{Conclusion}
The main innovation of \model is to jointly consider aspects and targets. It uses selective attention to model the relationships between target and context words, and aspects to attend to targeted contexts to predict sentiments. Users can ``query'' \model\ about sentiment of a particular aspect or target, or both. Our evaluation shows that \model\ outperforms state-of-the-art models on SemEval, SentiHood, and conflicting sentiment datasets. Our ablation study shows that jointly modeling aspects and targets with selective attention is superior to selective attention only, aspect attention only, and aspect with self-attention. 

\bibliographystyle{acl_natbib}
\bibliography{Zhe,chang}

\appendix
\section{Appendices}
\label{sec:appendix}
We present additional results here. 

\begin{table*}[!hbt]
\centering
\caption{Comparing performance of \model, BERT$_{\text{BASE}}$, and BERT$_{\text{LARGE}}$ on all tasks. Each experiment is repeated five times with different random seeds. Here, ``sd'' indicates one standard deviation.}
\label{tab:semeval1}
\scalebox{1} {
\begin{tabular}{l@{~~} r@{~~~~~}r@{~~~~~}r@{~~~~~}r@{~~~~~}}\toprule
Model & \multicolumn{2}{c}{{SemEval-14-A}} & \multicolumn{2}{c}{{SemEval-14-T}} \\\midrule
& Accuracy & $\text{F}_1$& Accuracy & $\text{F}_1$\\\midrule
BERT$_{\text{BASE}}$ & 84.16\sd0.53 & 76.37\sd0.84 & 79.19\sd0.62& 67.88\sd1.39\\
\model-BERT$_{\text{BASE}}$ & \textbf{84.53\sd0.13}& \textbf{77.15\sd0.48} & \textbf{82.93\sd0.57}& \textbf{74.96\sd1.02} \\
p-value & 0.16 & 0.11 & 9.04e-6 & 1.63e-05 \\\midrule
BERT$_{\text{LARGE}}$ & 85.17\sd0.55& 77.41\sd0.80 & 79.82\sd0.34 & 68.31\sd0.84\\
\model-BERT$_{\text{LARGE}}$ & 85.17\sd0.52& \textbf{77.46\sd0.77} & \textbf{83.30\sd0.25}& \textbf{74.96\sd0.25} \\
p-value & 1.00 & 0.91 & 8.21e-8 & 1.44e-7 \\\bottomrule \\ \toprule

Model & \multicolumn{2}{c}{{SemEval-15}} & \multicolumn{2}{c}{{SemEval-16}} \\\midrule
& Accuracy & $\text{F}_1$ & Accuracy & $\text{F}_1$\\\midrule
BERT$_{\text{BASE}}$ & 79.10\sd1.03 & 61.09\sd1.82 & 86.52\sd0.49 & 70.64\sd0.99\\
\model-BERT$_{\text{BASE}}$ & \textbf{83.15\sd1.04} & \textbf{64.83\sd2.49}& \textbf{89.31\sd0.56}& \textbf{75.33\sd1.46}\ \\
p-value & 2.63e-04 & 2.67e-02 & 3.04e-05 & 3.47e-4 \\\midrule
BERT$_{\text{LARGE}}$ & 83.81\sd0.64 & \textbf{65.41 \sd0.91} & 88.85\sd0.44 & 74.25\sd1.27\\
\model-BERT$_{\text{LARGE}}$ & \textbf{84.85\sd0.46} & 64.96\sd1.19 & \textbf{90.45\sd0.63} & \textbf{74.61\sd2.37} \\
p-value & 0.02 & 0.52 & 1.66e-3 & 0.77 \\\bottomrule \\ \toprule

Model & \multicolumn{2}{c}{{SentiHood-dev}} & \multicolumn{2}{c}{{SentiHood-test}} \\\midrule
& Accuracy & $\text{F}_1$& Accuracy & $\text{F}_1$\\\midrule
BERT$_{\text{BASE}}$ & 86.52\sd0.60 & 82.33\sd0.83 &86.54\sd0.47 & 82.63\sd0.93\\
\model-BERT$_{\text{BASE}}$ &\textbf{91.55\sd0.79} & \textbf{89.05\sd1.02}&\textbf{91.10\sd0.37} & \textbf{88.73\sd0.45} \\
p-value & 3.39e-06 & 3.08e-6 & 1.43e-7 & 1.08e-6 \\\midrule
BERT$_{\text{LARGE}}$ &87.38\sd0.35 &83.32\sd0.38&87.03\sd0.30&83.09\sd0.41\\
\model-BERT$_{\text{LARGE}}$ & \textbf{88.48\sd0.36} & \textbf{84.91\sd4.68} &\textbf{91.34\sd0.59}&\textbf{89.03\sd0.79} \\
p-value & 0.51 & 0.47 & 5.03e-7 & 3.94e-7 \\\bottomrule
\end{tabular}
}
\end{table*}

\clearpage
\begin{table*}[htb]
\centering
\caption{Comparing accuracy of \model and BERT on all tasks. Each experiment is repeated five times with different random seeds. ``\sd'' indicates one standard deviation. The p-value row indicates if the accuracy of BERT is significantly different from \model, measured by two sample $t$-test. We compare each of the 25 combination of experiments between BERT and \model. The ``BERT $\geq$ \model'' row counts the number of combinations where BERT is no worse than \model, and how many of them are significant, measured by McNemar test. Similarly, ``BERT $<$ \model'' counts the number of combinations where BERT is worse than \model. For example, on SemEval-14-A, BERT is no worse than \model in 12 combinations, none of which are significant. BERT performs worse than \model in the other 13 combinations, none of which are significant either.}
\label{tab:semeval_bert}
\scalebox{.84} {
\begin{tabular}{l@{~~} r@{~~~~~} r@{~~~~~} r@{~~~~~} r@{~~~~~} r@{~~~~~} r@{~~~~~}}\toprule
Model & SemEval-14-A & SemEval-14-T & SemEval-15 & SemEval-16 & SentiHood-dev & SentiHood-test \\\midrule
BERT & \textbf{86.15\sd0.52} & 80.39\sd0.63 & 83.72\sd0.96 & 88.52\sd0.49 &87.60\sd 0.17&87.09\sd 0.20\\
\textbf{\model} & {86.03\sd0.16} & \textbf{84.90\sd0.45} & \textbf{86.27\sd0.57} & \textbf{90.10\sd0.22} & \textbf{92.17\sd0.37} & \textbf{91.34\sd0.31}\\\midrule
p-value & 0.64 & 1.16e-6 & 9.22e-4 & 1.64e-4 & 8.32e-9 & 5.08e-9 \\
BERT $\geq$ \model & 12 (0) & 0 (0) & 0 (0) & 0 (0) & 0 (0) & 0 (0) \\
BERT $<$ \model & 13 (0) & 25 (25) & 25 (17) & 25 (8) & 25 (25) & 25 (25) \\
\bottomrule
\end{tabular}
}
\end{table*}

\begin{table*}[htb]
\centering
\caption{Comparing accuracy of \model model variants on SemEval-15. Each experiment is repeated five times with different random seeds. ``\sd'' indicates one standard deviation. The p-value column indicates if the accuracy of the variant is significantly different from \model, measured by two sample $t$-test. We compare each of the 25 combination of experiments between the variants and \model. The ``variant $\geq$ \model'' column counts the number of combinations where the variant is better than \model, and how many of them are significant, measured by McNemar test. Similarly, ``variant $<$ \model'' counts the number of combinations where the variant is worse than \model. For example, \model-Sel is better than \model in eight combinations but none of them are significant. \model is better than \model-Sel in 17 combinations where nine of them are significant.}
\label{tab:ablation_a}
\scalebox{1} {\begin{tabular}{lccrrrr}\toprule
 & & & & & \multicolumn{2}{c}{McNemar significance test} \\
Model & Aspect & Target & Accuracy & p-value & variant $\geq$ \model & variant $<$ \model \\\midrule
\model-Asp & Yes & -- &84.09\sd0.68 & 2.23e-8 & 0 (0) & 25 (14)\\
\model-Sel & -- & Selective &85.16\sd1.24 & 4.65e-6 & 8 (0) & 17 (9)\\
\model-Asp-Full & Yes & Self &82.01\sd1.95 & 5.72e-6 & 0 (0) & 25 (14) \\
\textbf{\model} & \textbf{Yes} & \textbf{Selective} &\textbf{86.27\sd0.57} & -- & -- & -- \\
\bottomrule
\end{tabular}
}
\end{table*}

\end{document}